\documentclass[lettersize,journal]{IEEEtran}
\usepackage{amsmath,amsfonts} 
\usepackage{amssymb}
\usepackage{algorithmicx}
\usepackage{algpseudocode}
\usepackage[linesnumbered,ruled]{algorithm2e}
\SetKwBlock{Init}{Initialize}{end}
\SetKwBlock{Train}{Train Model}{end}
\SetKwBlock{Extract}{Extract Action Effect Features}{end}
\SetKwBlock{Plan}{Plan Generation}{end}
\SetKwBlock{Parallel}{Parallel Executation}{end}
\usepackage{algpseudocode}
\usepackage[export]{adjustbox}
\usepackage{array} 
\usepackage{booktabs}
\usepackage{cite}
\usepackage{color}
\usepackage[dvipsnames]{xcolor}
\usepackage[hang,flushmargin]{footmisc}
\makeatletter
\newcommand{\algorithmfootnote}[2][\footnotesize]{%
	\let\old@algocf@finish\@algocf@finish
	\def\@algocf@finish{\old@algocf@finish
		\leavevmode\rlap{\begin{minipage}{\linewidth}
				#1#2
		\end{minipage}}%
	}%
}
\makeatother 
\usepackage{graphicx} 
\usepackage{hyperref}
\hypersetup{
	pagebackref=false,
	hypertex=true,
	colorlinks=true,
	linkcolor=black,
	anchorcolor=black,
	citecolor=blue,
	urlcolor=black
}
\usepackage{bm}
\usepackage{makecell}
\usepackage{multirow}
\usepackage{pdfpages}
\usepackage{pifont}
\usepackage{stfloats} 
\usepackage[caption=false,font=footnotesize,labelfont=rm,textfont=rm]{subfig}
\usepackage{subfig}
\usepackage{textcomp} 
\usepackage{threeparttable}
\usepackage{url} 
\usepackage{utfsym}
\usepackage{verbatim} 
\begin{document}
\title{Task-Agnostic Learning to Accomplish New Tasks}
\author{
	Xianqi Zhang,
	Xingtao Wang,
	Xu Liu,
	Wenrui Wang,
	Xiaopeng Fan,~\IEEEmembership{Senior Member,~IEEE},\\
	and Debin Zhao,~\IEEEmembership{Member,~IEEE}

	\thanks{
		X. Zhang, X. Wang, X. Liu, W. Wang, X. Fan, and D. Zhao are with the 
		Faculty of Computing, Harbin Institute of Technologyy, Harbin 150001, China. 
		X. Fan are also with the
		Peng Cheng Laboratory, Shenzhen 518000, China,
		and 
		Harbin Institute of Technology Suzhou Research Institute, Suzhou 215000, China.
		(E-mail:
		zhangxianqi@stu.hit.edu.cn; 
		xtwang@hit.edu.cn;
		20B903008@stu.hit.edu.cn;
		21B303001@stu.hit.edu.cn;
		fxp@hit.edu.cn; 
		dbzhao@hit.edu.cn.)
	}
	\thanks{
		\textit{Corresponding author: Xingtao Wang.}
	}
}

\markboth{IEEE Transactions on Cognitive and Developmental Systems}%
{Shell \MakeLowercase{\textit{et al.}}: A Sample Article Using IEEEtran.cls for IEEE Journals}


\maketitle

\begin{abstract}
Reinforcement Learning (RL) and Imitation Learning (IL) have made great progress in robotic decision-making in recent years. 
However, these methods show obvious deterioration for new tasks that need to be completed through new combinations of actions. 
RL methods suffer from reward functions and distribution shifts, while IL methods are limited by expert demonstrations which do not cover new tasks. 
In contrast, humans can easily complete these tasks with the fragmented knowledge learned from task-agnostic experience. 
Inspired by this observation, this paper proposes a task-agnostic learning method (TAL for short) that can learn fragmented knowledge only from task-agnostic data to accomplish new tasks. 
TAL consists of four stages. 
First, the task-agnostic exploration is performed to collect data from interactions with the environment.
The collected data is organized via a knowledge graph.
Second, an action feature extractor is proposed and trained using the collected knowledge graph data for task-agnostic fragmented knowledge learning. 
Third, a candidate action generator is designed, which applies the action feature extractor on a new task to generate multiple candidate action sets. 
Finally, an action proposal network is designed to produce the probabilities for actions in a new task according to the environmental information. 
The probabilities are then used to generate order information for selecting actions to be executed from multiple candidate action sets to form the plan.
Experiments on a virtual indoor scene show that the proposed method outperforms the state-of-the-art offline RL methods and IL methods by more than 20\%.
\end{abstract}


\section{Introduction}
\label{section_1}
\IEEEPARstart{A} robot is expected to learn and work like humans. 
At present, the robot usually uses sensors\cite{tcds_2_1, tcds_2_3, tcds_2_5} to perceive environmental information and uses planning and control algorithms to make decisions \cite{tcds_3_1, tcds_3_2, tcds_3_3}.
For tasks that require decision-making, there are always new tasks that must be completed through new combinations of actions. 
For example, ``prepare for reading'' needs ``turn on the light and push the chair near the table'' to perform. 
Actions such as ``turn on the light'' and ``push the chair near the table'' may have been done before independently, but not sequentially together. 
Humans can easily complete these new tasks, but it is very challenging for Reinforcement Learning (RL)\cite{wang2022deep} and Imitation Learning (IL)\cite{hussein2017imitation} based decision-making methods.

RL and IL are two main paradigms in robotic decision-making.
RL refers to learning from interactions with the environment and has made significant advances in robotics applications \cite{singh2021reinforcement, van2016deep, keneshloo2019deep, schulman2017proximal, zhang2021safe, tcds_1_1, tcds_1_2, tcds_3_4, tcds_3_5, tcds_3_6}.
RL policies are typically driven by reward signals \cite{sutton2018rl, hadfield2017inverse, he2022assisted}, enabling them to learn and output appropriate actions to complete the task.
However, over-reliance on task-specific reward signals can negatively impact generalization performance.
Moreover, when encountering new tasks, the distribution shift problem \cite{kumar2020conservative, yu2021combo} may be more serious, further complicating their performance.
Different from RL, IL refers to learning from demonstrations \cite{osa2018algorithmic, finn2016guided, ho2016generative, torabi2018behavioral}. 
IL methods need to collect expert data in advance and teach the agent to imitate the expert behavior to achieve the goal \cite{osa2018algorithmic}. 
However, these methods are usually limited by expert demonstrations, that is, their performance tends to degrade when expert demonstrations do not cover new tasks.

In contrast, humans can easily complete these new tasks with the fragmented knowledge learned from task-agnostic experience. 
An important way for humans to learn fragmented knowledge is to learn from task-agnostic interactions with environments.
For example, when a child interacts with objects in the environment without a specific goal in mind, he/she may learn fragmented knowledge such as ``a block can be picked up'' or ``some blocks can be combined in a certain way.'' 
To complete a new task, a child may first roughly estimate the actions required to complete the task according to the fragmented knowledge, and then determine the execution order of the actions based on the environmental state. 
For example, to construct a desired shape, a child can estimate the required blocks and then combine them in a certain order.

In this paper, inspired by how humans handle new tasks,
i.e., learn the fragmented knowledge $\! \rightarrow \! $ select actions that will be used for the task $\! \rightarrow \! $ decide the action execution order,
we propose a task-agnostic learning method (TAL for short) that can learn fragmented knowledge from task-agnostic data to accomplish new tasks.
TAL contains four stages: task-agnostic environment exploration, action feature extraction, candidate action generation, and plan generation by action proposal network. 
Compared with RL and IL methods, TAL alleviates the limitations of reward functions and expert demonstrations.

In previous studies, ``Task-Agnostic" has multiple referential meanings.
Two related types are introduced here.
1) The first type is (task-agnostic) meta-learning \cite{vettoruzzo2024advances} and (task-agnostic) continual learning \cite{wang2024comprehensive}, which means the learning methods are not limited to research areas or specific types of tasks, such as classification and detection in computer vision.
Meta-learning aims to learn how to adapt quickly to new tasks. 
Continual learning aims to learn from a sequence of experiences continually and avoid problems such as catastrophic forgetting. 
Compared to meta-learning, our method does not fine-tune on new tasks or generate network parameters according to sample information. 
Compared to continual learning, our method does not learn from experiences one by one, but learns knowledge from batches of data.
2) The second type is task-agnostic (or reward-free, intrinsic reward-based) exploration in RL \cite{amin2021survey, jin2020reward, mutti2021task, zhang2020task, parisi2021interesting}. 
Previous studies usually focus on how to use task-agnostic intrinsic rewards to guide the agent to fully explore the environment. 
The exploration data is usually combined with a task-specific reward function to train a policy, or to learn knowledge for better exploration.
Zhang \textit{et al.} \cite{zhang2020task} use multiple reward functions to augment reward-free exploration data and then train the policy to solve multiple tasks. 
Parisi \textit{et al.} \cite{parisi2021interesting} use task-agnostic exploration data and intrinsic rewards to train a state-value function, and then use the function as a bias of the policy to explore new environments.
Compared to these methods, our method does not use reward functions or require additional environment interactions for fine-tuning on new tasks.

In this paper, we refer to the pattern in which the model learns task-agnostic knowledge only from task-agnostic data as Task-Agnostic Learning.
This work explores decision-making and learning methods that minimize human involvement, allowing agents to learn only from task-agnostic data and leverage the learned fragmented knowledge to solve new tasks. 
This approach aims to create more autonomous systems that generalize across tasks without extensive task-specific training or human supervision, resulting in more efficient and adaptable robots.

The main contributions of this paper are summarized as follows:
\begin{itemize}
	\item{A task-agnostic learning method (TAL for short) is proposed, which can learn fragmented knowledge from task-agnostic data to accomplish new tasks.}
	\item{An action feature extractor is proposed and trained using the task-agnostic exploration data for fragmented knowledge learning.}
	\item{A candidate action generator is proposed, which applies the action feature extractor on a new task to generate multiple candidate action sets.}
	\item{An action proposal network is designed to generate execution order information for actions in multiple candidate action sets according to the environmental information. 
	}
\end{itemize}

The rest of this paper is organized as follows. 
Related works are briefly reviewed in Section \ref{section_2}. 
The proposed framework is described in Section \ref{section_3}. 
The experimental results are presented in Section \ref{section_4}. 
Section \ref{section_5} provides the conclusion.

\section{Related Work}
\label{section_2}

\subsection{Reinforcement Learning}
Reinforcement Learning (RL) has been widely used in robotics-related scenarios \cite{singh2021reinforcement, van2016deep, keneshloo2019deep, schulman2017proximal, zhang2021safe}. 
Recently,
goal-conditioned RL is proposed to enable an agent to be able to perform multiple tasks \cite{nair2020goal, campero2020learning, mendonca2021discovering, liu2022goal, mezghani2022walk, chevalier2018minimalistic, gupta2020relay}. 
Different from traditional RL methods \cite{mnih2015human, lillicrap2015continuous, fujimoto2019off}, 
the agent in goal-conditioned RL is anticipated to consider both task and environmental information when making decisions \cite{liu2022goal, mezghani2022walk}. 
The generalization of RL has always attracted much attention and is often significantly affected by factors such as reward functions and data distribution shifts.

Reward functions play a crucial role in RL as it is the primary basis for altering the policy \cite{sutton2018rl}. 
Hadfield \textit{et al.} \cite{hadfield2017inverse} introduce the approximate method for solving the inverse reward design problem. 
Devidze \textit{et al.} \cite{devidze2021explicable} propose a new framework to investigate the explicable reward function design from the perspective of discrete optimization. 
He \textit{et al.} \cite{he2022assisted} propose an assisted reward design method that accelerates the design process by anticipating and influencing future design iterations.

Many works aim to alleviate the data distributional shift problem \cite{levine2020offline}, i.e., the agent cannot perform well when training and testing data differ significantly. 
Kumar \textit{et al.} \cite{kumar2020conservative} propose Conservative Q-Learning (CQL), where they used a Q-value regularizer to constrain the learned Q-function. 
Yu \textit{et al.} \cite{yu2021combo} propose Conservative Offline Model-Based Policy Optimization (COMBO) to learn a conservative Q-function by penalizing the value function on out-of-support state-action tuples. 
Wiles \textit{et al.} \cite{wiles2021fine} analyse the distributional shift problem in detail and gave some suggestions, such as data augmentation and pre-training.

RL methods that heavily rely on task-specific rewards often struggle to generalize to new tasks. 
When handling new tasks, the distribution shift problem may become more severe, further hindering the generalization of RL methods.
Additionally, since our goal is to enable the agent to learn directly from task-agnostic data without a specific task, designing an appropriate reward function for RL becomes challenging.
Moreover, given the uncertainty of tasks, all fragmented knowledge is valuable. 
Since RL typically learns through trial and error, using a task-specific reward function to distinguish between valid and invalid knowledge is unsuitable.

\subsection{Imitation Learning}
Imitation Learning (IL), also known as learning from demonstration, refers to making a robot imitate the behavior of experts \cite{hussein2017imitation, torabi2019recent}.

A branch of the IL is Inverse Reinforcement Learning (IRL), which seeks to recover reward function from demonstrations. 
Finn \textit{et al.} \cite{finn2016guided} employ neural networks for learning cost function and combined IRL to teach the agent to carry out identical activities as the expert. 
Ho \textit{et al.} \cite{ho2016generative} propose a new framework named Generative Adversarial Imitation Learning (GAIL), which combined IRL with the idea of generative adversarial, and received widespread attention \cite{finn2016connection, merel2017learning, wu2019imitation, garg2021iq, cao2021learning}.

Another branch is Behavioral Cloning (BC), which aims to mimic expert behavior through supervised learning. 
Recently, the combination of BC and deep learning has attracted extensive attention \cite{finn2017deep, agrawal2016learning, edwards2019imitating, ehsani2020use}. 
Sharma \textit{et al.} \cite{sharma2019third} train two models, in which the high-level model generated a series of first-person sub-goals based on the video from the third-person perspective, and the low-level model predicted the actions necessary to fulfill the sub-goals. 
This paradigm is similar to hierarchical RL \cite{yang2018hierarchical, dayan1992feudal, parr1997reinforcement, sutton1999between, dietterich2000hierarchical, vezhnevets2017feudal}. 
In addition, to alleviate the difficulty of collecting expert demonstrations, Lynch \textit{et al.} \cite{lynch2020learning} let humans manipulate robots to collect task-agnostic data, referring to it as play data. 
Since there is no fixed goal, humans operate according to their curiosity. The meaningful action sequences collected in this way can be viewed as different skills. 
Play sequences are first sampled from the play data and encoded into the latent plan space. 
Then, a goal-conditioned policy was trained to complete multiple tasks. 
This work achieves good performance, but it is still limited to human control and the patterns of goal-conditioned IL.

For IL methods, collecting large amounts of expert demonstrations is costly, and performance often deteriorates if expert demonstrations do not cover new tasks.
Since our goal is to enable the agent to learn solely from task-independent data, i.e., there are no expert data corresponding to new tasks, IL methods become ineffective.

\subsection{Meta-Learning}
Meta-learning, also known as learning to learn, has been a very popular research direction in recent years.
Meta-learning aims to train a network to adapt quickly to new tasks.
Specifically, it learns how to learn tasks during the meta-training phase so that new tasks can be effectively learned during meta-testing \cite{vettoruzzo2024advances}.
Generally, meta-learning methods can be divided into three categories \cite{vettoruzzo2024advances, hospedales2021meta}:
1) Optimization-based methods aim to learn meta-parameters that can quickly adapt to new tasks.
Chelsea Finn \textit{et al.} \cite{finn2017model} propose MAML to learn the initial parameters of a network, which can be quickly optimized with gradient descent on new tasks.
2) Black Box/Model-based methods train a block-box network $f_{\theta}$ to predict parameters $\phi$ of the network $h_{\phi}$. 
During the meta-testing phase, $h_{\phi}$ is used to make predictions on the testing tasks.
Such methods usually only output a portion of the parameters of the prediction network, for example, Wang \textit{et al.} \cite{wang2024hyper} customize a part of the parameters of the mesh denoising network with a meta-network.
3) Metric/Non-parametric-based methods use a learned metric to compare test data with training data in the embedding space and assign the label of matching training data \cite{chen2020variational}.
Recently, meta-RL has been proposed to combine the advantages of meta-learning and RL \cite{beck2023survey}. 
As a special case of meta-learning, meta-RL aims to improve the generalization performance of the policies.
Like meta-learning, meta-RL also includes optimization-based \cite{mitchell2021offline}, model-based \cite{lin2020model}, and metric-based \cite{li2020focal} methods.

Compared with meta-learning, our method does not fine-tune on new tasks (optimization-based), 
or generate different network parameters based on task information (black box/model-based), 
or learn a metric to compare the test task and the training task (metric/non-parametric-based). 
Instead, our method learns knowledge only based on task-agnostic exploration data and directly uses the learned knowledge to complete new tasks.

\subsection{Continual Learning}
Continual learning (CL) is an important way to solve the generalization problem of intelligent agents when data distribution changes. 
Recently, many works have been proposed to deal with the main challenges of CL, such as catastrophic forgetting, memory stability, and learning plasticity \cite{wang2024comprehensive}. 
There are many types of CL methods, such as regularization-based, optimization-based, memory/replay-based, and so on. 
An important difference between CL and meta-learning is that CL learns from a sequence of experiences where all data is not available at once \cite{lesort2020continual}, while meta-learning mainly focuses on scenarios where a batch of training tasks is available \cite{vettoruzzo2024advances}. 
There are also some works combining CL and meta-learning \cite{beaulieu2020learning, gupta2020look}.
Same as meta-RL, some works combine CL and RL as CRL, which enables the RL agent to continue learning knowledge \cite{abel2024definition, khetarpal2022towards, caccia2023task}.

Although CL is an effective way to improve the generalization and transfer capabilities of models, it is quite different from our method. 
On the one hand, our method does not learn from sequence experiences one by one, but learns knowledge from batch data. 
On the other hand, compared with problems in CL, such as catastrophic forgetting, we emphasize knowledge learning from task-agnostic data, as well as knowledge retrieval and combination for addressing new tasks.

\begin{figure*}[!t]
	\centering
	\includegraphics[width=.92\linewidth]{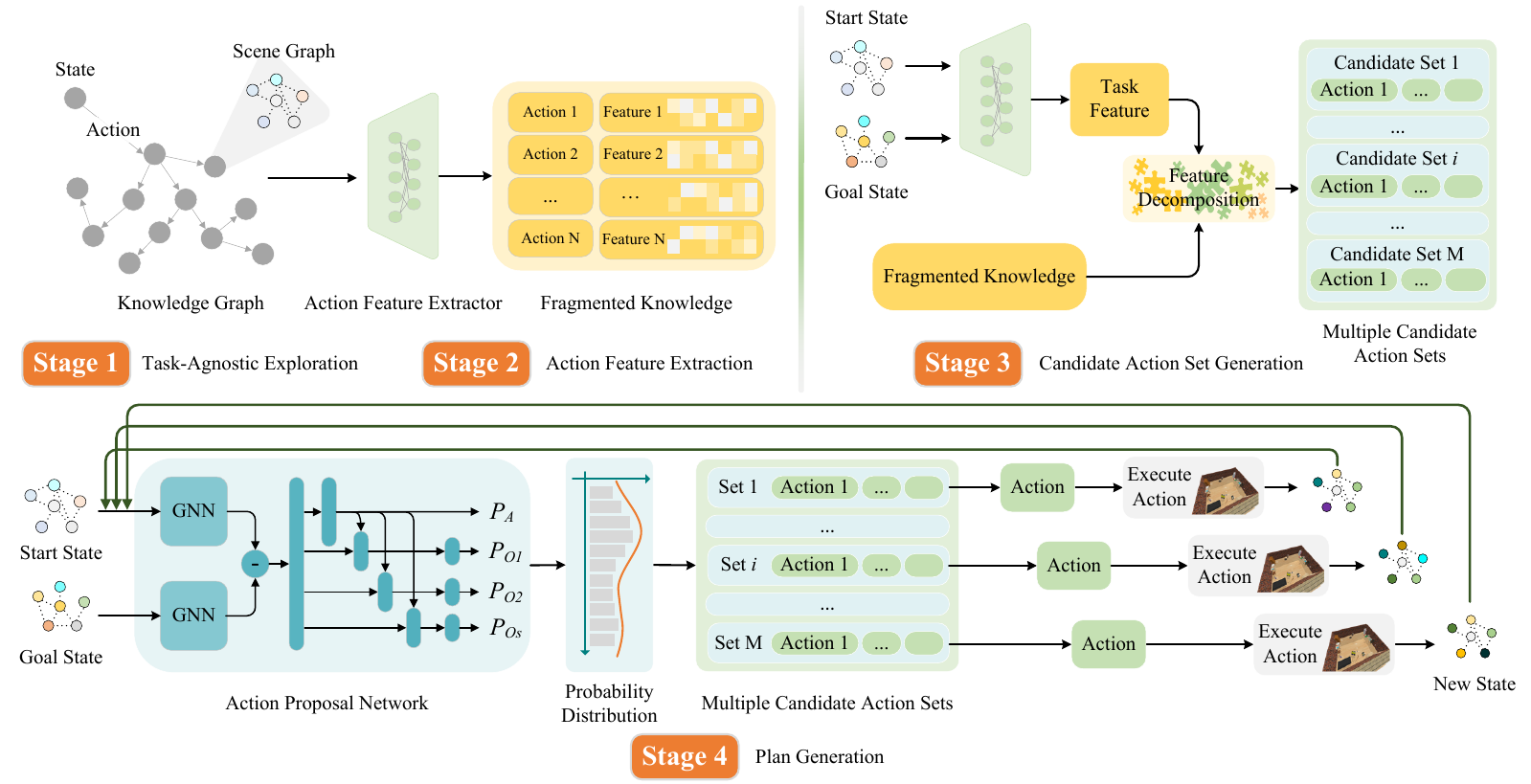}
	\caption{The framework of TAL.
	}
	\label{fig_pipeline}
\end{figure*}

\section{Framework}
\label{section_3}
The proposed TAL contains four stages.
First, the task-agnostic exploration is performed to collect data from interactions with the environment. The collected data is organized via a knowledge graph.
Second, an action feature extractor is proposed to learn task-agnostic fragmented knowledge.
Third, a candidate action generator is designed to generate multiple candidate action sets for a new task.
Finally, an action proposal network is designed to generate the plan.
The framework of TAL is shown in Fig.~\ref{fig_pipeline}.
In the following subsections, we will elaborate on the four stages respectively.

\subsection{Task-Agnostic Exploration}
\label{section_exploration}
The task-agnostic exploration is performed to collect data from interactions with the environment. 
The collected data is organized via a knowledge graph.

\subsubsection{Knowledge Graph}
A knowledge graph: $\mathcal{G} = (\mathbb S, \mathbb A)$ is built during environment exploration to collect data, where $\mathbb S$ is the set of nodes and $\mathbb A$ is the set of directed edges. 
A node $s_i \in \mathbb S$ represents an environmental state. 
An edge $(s_i, s_j)$ is created if an action $a_{ij}$ can be executed.

Previous works \cite{pathak2017curiosity, burda2018exploration, badia2019never, sekar2020planning} typically employ a sequential structure to organize the explored data. 
Since the environment is always initialized to the same or similar state, the sequence data may contain redundant fragments, making the exploration difficult. 
In contrast, we build a knowledge graph to organize the explored data. 
With the ability to restore the environment to various earlier states, the knowledge graph is more compact and makes it easier to explore the environment.

\subsubsection{Exploration}
At the start of the task-agnostic exploration, the node corresponding to the initial state of the environment is created, while the edge is an empty set. 
A round of exploration begins by randomly selecting a node $s_i$ from $\mathbb S$. 
An action is then randomly sampled from the set of all executable actions. 
If the execution of the action fails, another action is sampled for execution. 
If the action can be executed successfully, a new node $s_j$ is created to correspond to the new environmental state, and the edge from $s_i$ to $s_j$ is created. 
The next step of exploration starts from $s_j$. 
(The subscripts $i$ and $j$ are used to indicate different states.)
A round of exploration consists of several step explorations. 
Numerous rounds of exploration are carried out to sufficiently explore the environment. 
Finally, a knowledge graph with extensive data is built. 
The task-agnostic exploration is shown in Algorithm \ref{algorithm_explore}.
More details on using discrete actions for environment-specific exploration and dataset generation will be introduced in Section \ref{section_VI_A_setup}.

\subsection{Action Feature Extraction}
\label{subsection_AFE}
In this subsection, an Action Feature Extractor (AFE) is proposed to learn task-agnostic fragmented knowledge from the collected knowledge graph data. 
The action feature extraction is based on the states before and after an action is executed, which can be expressed as follows:
\begin{equation}\label{eq1}
	\begin{aligned}
		\boldsymbol{F}_{ij}={\rm AFE}(s_i,\ s_j\ |\ a_{ij}).
	\end{aligned}
\end{equation}
$\boldsymbol{F}_{ij}$ is the action feature. $s_i$ and $s_j$ are environmental states before and after the action $a_{ij}$ is executed.
Next, we first introduce the structure of AFE and then elaborate on the training strategy.

\subsubsection{Structure}
\begin{figure}[!t]
	\centering
	\includegraphics[width=3.4in]{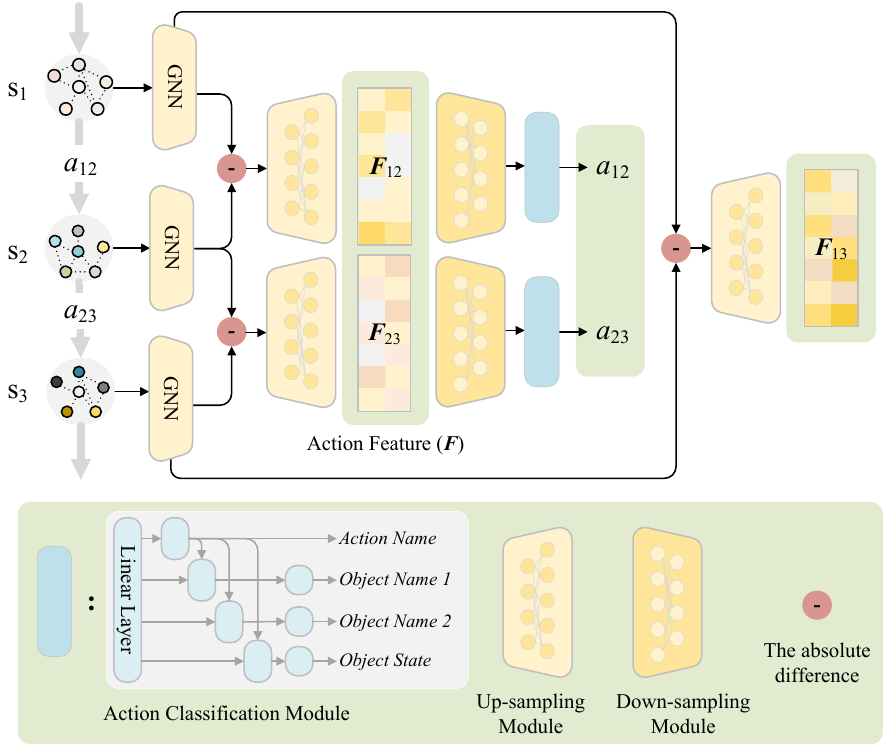}
	\caption{The action feature extractor.}
	\label{fig_act_effect}
\end{figure}

The structure of AFE is shown in Fig. \ref{fig_act_effect}. 
AFE takes the states before and after an action is executed as input, and outputs the action feature. 
AFE consists of two Graph Neural Networks (GNNs) and an up-sampling module. 
The GNNs are implemented with 3 Gated-GCN layers, using ${\rm Tanh}$ as the activation function and a hidden dimension of 128. 
The up-sampling module is implemented with linear layers, using ${\rm ReLU}$ as the activation function.
Each GNN captures one state feature from one input state. 
The absolute difference of the two captured state features is fed into the up-sampling module to extract the action feature.
The action feature extracted by the up-sampling module is in a higher dimension space, so the features of different actions are more discriminative.

\subsubsection{Training Strategy}
In the following, we explain the training strategy in terms of training data and loss functions.

We sample paths of various lengths from the knowledge graph. 
The nodes (states) and edges (actions) in each path are extracted to form a trajectory. 
Trajectories are used to generate training samples. 
A training sample consists of three consecutive nodes and two edges between them. 
Next, we take a training sample $[s_1, a_{12}, s_2, a_{23}, s_3]$ as an example to introduce the training strategy. 
An action $a_{ij}$ is represented by action name and parameters (object names and object state).

Due to the lack of supervised information for the training of AFE, an action classifier is introduced. 
The classification labels supervise the training of the classifier and AFE. 
The action classifier can be represented as
\begin{equation}\label{eq2}
	\begin{aligned}
		\boldsymbol{T}_{ij}&={\rm Classifier}(\boldsymbol{F}_{ij}).\\
	\end{aligned}
\end{equation}
$\boldsymbol{T}_{ij}$ represents the predicted action tensor. 
$\boldsymbol{F}_{ij}$ is the action feature.
The action classifier consists of a down-sampling module and an action classification module, both of which are implemented by multiple linear layers.

AFE and the classifier are supervised by three loss functions.
\begin{equation}\label{eq3}
	\begin{aligned}
		\mathcal{L}_{AFE}=\mathcal{L}_{cls}+\mathcal{L}_{f}+\mathcal{L}_{add}.
	\end{aligned}
\end{equation}
The action classification loss ($\mathcal{L}_{cls}$) ensures that extracted features are associated with actions. 
The feature distinguish loss ($\mathcal{L}_{f}$) constrains the features of different actions to be more discriminative. 
The additivity loss ($\mathcal{L}_{add}$) ensures the additive property of the action feature space.

$\mathcal{L}_{cls}$ is calculated according to predictions of the classifier and action classification labels. 
For the training sample $[s_1, a_{12},$ $s_2, a_{23}, s_3]$, $\mathcal{L}_{cls}$ is the sum of the prediction errors for the two actions.
\begin{equation}\label{eq4}
	\begin{aligned}
		\mathcal{L}_{cls}=
		\mathcal{L}_{bce}(\boldsymbol{T}_{\! 12},\ \boldsymbol{T}_{\! 12}^\ast)
		+\mathcal{L}_{bce}(\boldsymbol{T}_{\! 23},\ \boldsymbol{T}_{\! 23}^\ast).
	\end{aligned}
\end{equation}
Here, $\boldsymbol{T}_{\! 12}$ and $\boldsymbol{T}_{\! 23}$ are predicted action tensors of $a_{12}$ and $a_{23}$ respectively. 
$\boldsymbol{T}_{\! 12}^*$ and $\boldsymbol{T}_{\! 23}^*$ represent the ground truth tensors of $a_{12}$ and $a_{23}$. 
In the adopted environment, an action consists of an action name and one or two parameters (object name or object state), so $\boldsymbol{T}^*$ is the concatenation of one-hot embeddings of the action name and parameters.
$\mathcal{L}_{bce}$ represents the binary cross-entropy loss. 
Minimizing $\mathcal{L}_{cls}$ ensures that extracted features are associated with actions.

$\mathcal{L}_{f}$ is computed by the features of two actions. 
The loss function reduces the feature distance of the same action while increasing the distance between different actions.
\begin{equation}\label{eq5}
	\begin{aligned}
		\mathcal{L}_{f} \! = \!
		\left\{
		\begin{array}{lr}\!\!
			1-\cos(\boldsymbol{F}_{\! 12},\ \boldsymbol{F}_{\! 23}), \: \: \: \: \: \: \: \: \: \: \: \: \: \: \: \: \: \: if \ a_{12}=a_{23}\\
			\!\!\max \big(0,\cos(\boldsymbol{F}_{\! 12},\ \boldsymbol{F}_{\! 23})-\varepsilon \big), \: if\ a_{12} \neq a_{23}.\\
		\end{array}\right.
	\end{aligned}
\end{equation}
$\boldsymbol{F}_{\! 12}$ and $\boldsymbol{F}_{\! 23}$ are the action features of $a_{12}$ and $a_{23}$ respectively. $\varepsilon$ is a small offset. 
Minimizing $\mathcal{L}_{f}$ constrains the features of different actions to be more discriminative.

$\mathcal{L}_{add}$ is calculated according to three action features. 
\begin{equation}\label{eq6}
	\begin{aligned}
		\mathcal{L}_{add}=
		\Vert (\boldsymbol{F}_{\! 12}+\boldsymbol{F}_{\! 23}) -  \boldsymbol{F}_{\! 13}\Vert_2^2.
	\end{aligned}
\end{equation}
Minimizing $\mathcal{L}_{add}$ ensures the additive property of the action feature space.

AFE is trained end to end using the loss function $\mathcal{L}_{AFE}$.

\begin{algorithm}[t]
	\DontPrintSemicolon
	\SetAlgoLined
	\caption{Task-agnostic exploration}
	\label{algorithm_explore}
	\KwOut {A knowledge graph $\mathcal{G} = (\mathbb S, \mathbb A)$}
	\Init{
		$\mathbb A_{all}$ \textcolor{gray}{// The set of all executable actions.}\\
		$\mathbb A = \emptyset$ \textcolor{gray}{// The set of edges that represent actions.}\\
		$\mathbb S = \{s_0\}$ \textcolor{gray}{// $s_0$ is the initial environmental state.}\\
	}
	\For {$round=0 \;\; to \;\; max\_round$} {
		$s_i$ = random\_select\_node($\mathbb S$)\\
		$attempt = 0$\\
		\For{$step=0 \;\; to \;\; max\_step$} {
			$a$ = random\_select\_action($\mathbb A_{all}$)\\
			$success$ = execute\_action($a$, $s_i$)\\
			\While {$!success \; \& \;  attempt < max\_attempt$ } {
				$attempt = attempt + 1$\\
				$success$ = execute\_action($a$, $s_i$)\\
			}
			\If{$attempt >= max\_attempt$}{break}
			$s_j$ = new\_state\_node()\\
			$\mathbb S = \mathbb S \cup \{s_j\}$\\
			$a_{ij}$ = create\_edge($s_i$, $s_j$, $a$)\\
			$\mathbb A = \mathbb A \cup \{a_{ij}\}$\\
			$s_i \gets s_j$ \textcolor{gray}{// Next step starts from $s_j$.}\\
		}
	}
\end{algorithm}

\subsection{Candidate Action Generation}
In this subsection, a Candidate Action Generator (CAG) is proposed to generate a Candidate Action Set (CAS).

For a new task $T$, representing the current environmental state as $s_c$ and the goal state as $s_g$, the task feature $\boldsymbol{F}_{\! T}$ is extracted by the Action Feature Extractor (AFE).

\begin{equation}\label{eq7}
	\begin{aligned}
		\boldsymbol{F}_{\! T}={\rm AFE} (s_c,\ s_g).
	\end{aligned}
\end{equation}

The task feature reflects the environmental evolution, which is the superposition of the impacts of multiple actions taken to complete the task. 
According to the additive property of the feature space, the task feature can be decomposed into several action features, which can be expressed as:
\begin{equation}\label{eq8}
	\begin{aligned}
		\boldsymbol{I}_{\! A} \boldsymbol{F}_{\!\! A} = \boldsymbol{F}_{\! T}.
	\end{aligned}
\end{equation}
Here, $\boldsymbol{I}_{\! A} \in \{0, 1\}^n $ is the action index corresponding to the actions in the action set $\mathbb{A}_n$.  
$\mathbb{A}_n=\{a_1, a_2, ..., a_n\}$, $a_i$ represents an action and $n$ is the number of  actions.
$\boldsymbol{F}_{\!\! A} = \left[\boldsymbol{F}_1^T, \boldsymbol{F}_2^T, … , \boldsymbol{F}_{\! n}^T\right]^{T}$ is the features of actions in $\mathbb{A}_n$.

However, the action features are usually contaminated by noise, making Eq. (\ref{eq8}) only approximately true. 
\begin{equation}\label{eq9}
	\begin{aligned}
		\boldsymbol{\widetilde{I}}_{\! A} \boldsymbol{F}_{\!\! A} = \boldsymbol{F}_{\! T}.
	\end{aligned}
\end{equation}
$\boldsymbol{\widetilde{I}}_{\! A}$ is an approximation of $\boldsymbol{I}_{\! A}$.

To alleviate the impact of the noise, the Principal Component Analysis (PCA) is used.
\begin{equation}\label{eq10}
	\begin{aligned}
		\boldsymbol{F}_{\!\! A} = \boldsymbol{U} \rm diag(\boldsymbol{S}) \boldsymbol{V}^T.
	\end{aligned}
\end{equation}
Then $\boldsymbol{F}_{\!\! A}$ and $\boldsymbol{F}_{\! T}$ are mapped into another feature space.
\begin{equation}\label{eq11}
	\begin{aligned}
		\boldsymbol{\widetilde{I}}_{\! A} \boldsymbol{F}_{\!\! A} \boldsymbol{V} = \boldsymbol{F}_{\! T} \boldsymbol{V}.
	\end{aligned}
\end{equation}
Finally, the action index is obtained as:
\begin{equation}\label{eq12}
	\begin{aligned}
		\boldsymbol{\widetilde{I}}_{\! A} = (\boldsymbol{F}_{\! T} \boldsymbol{V})(\boldsymbol{F}_{\!\! A} \boldsymbol{V})^+.
	\end{aligned}
\end{equation}
Here, superscript $+$ means the Moore-Penrose pseudo-inverse.

As $\boldsymbol{\widetilde{I}}_{\! A}$ is an approximation of $\boldsymbol{I}_{\! A}$, each element in $\boldsymbol{\widetilde{I}}_{\! A}$ can be taken as the probability that an action will be used in the task. 
The actions corresponding to the $top\raisebox{0mm}{-}i$ values of $\boldsymbol{\widetilde{I}}_{\! A}$ are selected to form a CAS, denoted as $\mathbb{A}_i$ ($\mathbb{A}_i \subseteq \mathbb{A}_n$).

For a new task, on the one hand, $\mathbb{A}_i$ must contain as many correct actions as possible to solve the task. 
On the other hand, $\mathbb{A}_i$ should contain as few incorrect actions as possible. 
Since an agent does not know how many acts are required when solving a new task, the Multiple Candidate Action Sets (MCAS) are generated. 
Each CAS in MCAS contains a different number of actions, denoted as $\mathbb{M}=\{ \mathbb{A}_{i}, \mathbb{A}_{j}, ..., \mathbb{A}_{k}\}$.

\subsection{Plan Generation by Action Proposal Network}
\label{section_3_APN}
This subsection generates the final plan to accomplish a new task. 
First, an action proposal network is designed to generate order information by producing the probabilities for actions according to the environmental information. 
Second, we generate the plan based on the probabilities.

\subsubsection{Action Proposal Network} 
An Action Proposal Network (APN) is designed to generate the probability distributions for all executable actions, thereby providing action execution order information to a Candidate Action Set (CAS).
APN can be represented by
\begin{equation}\label{eq13}
	\begin{aligned}
		\boldsymbol{P}={\rm APN}(s_c,\ s_g).
	\end{aligned}
\end{equation}
$\boldsymbol{P}$ consists of four probability distributions, corresponding to action name, object name 1, object name 2, and object state, as shown in Table \ref{tab_env}.
Since the parameters (object name or state) are related to action name, similar to \cite{tuli2021tango}, APN first estimates action name and then the parameters.
APN is formulated as follows:
\begin{equation}\label{eq14}
	\begin{aligned}
		\boldsymbol{P}\: \ &={\rm concat}(\boldsymbol{P}_{\!\! A},\ \boldsymbol{P}_{\! O_1},\ \boldsymbol{P}_{\! O_2},\ \boldsymbol{P}_{\! O_s})\\
		\boldsymbol{P}_{\! A}\ &={\rm Softmax}\big({\rm MLP}_{\! A}(\boldsymbol{h}_1)\big)\\
		\boldsymbol{P}_{\! O_i}&={\rm Softmax}\big({\rm MLP}_{\! O_i}(\boldsymbol{h}_2)\big),\ \ i=1,\ 2,\ s.\\
	\end{aligned}
\end{equation}
Here, $\boldsymbol{h}_1\!=\!{\rm ReLU}\big({\rm MLP}_{\! h}(|{\rm GNN}(s_c) \! -\! {\rm GNN}(s_g)|)\big), \boldsymbol{h}_2\! =\!{\rm concat}(\boldsymbol{P}_{\!\! A},  \boldsymbol{h}_1)$.
${\rm Softmax}$ and ${\rm ReLU}$ are activation functions.
${\rm MLP}$ is a 3-layer perceptron.
${\rm GNN}$ is implemented with 3 Gated-GCN layers, using ${\rm Tanh}$ as the activation function and a hidden dimension of 128.
The GNNs used by $s_c$ and $s_g$ share weights. 
$\boldsymbol{P}_{\!\! A}$ is the predicted probabilities for action name, $\boldsymbol{P}_{\! O_1}$ and $\boldsymbol{P}_{\! O_2}$ are the predicted probabilities for object name 1 and object name 2, and $\boldsymbol{P}_{\! O_s}$ is the predicted probability for object state.

For an action in all executable actions, its probability is obtained by adding the four probabilities in $\boldsymbol{P}$.

A binary cross-entropy loss is used to train APN. 
The label $\boldsymbol{P}^\ast$ consists of four parts, corresponding to $\boldsymbol{P}$.
\begin{equation}\label{eq15}
	\begin{aligned}
		\boldsymbol{P}^\ast ={\rm concat}(\boldsymbol{P}_{\!\! A}^\ast,\ \boldsymbol{P}_{\! O_1}^\ast,\ \boldsymbol{P}_{\! O_2}^\ast,\ \boldsymbol{P}_{\! O_s}^\ast),\\
	\end{aligned}
\end{equation}
where ($\boldsymbol{P}_{\!\! A}^\ast,\boldsymbol{P}_{\! O_1}^\ast,\boldsymbol{P}_{\! O_2}^\ast,\boldsymbol{P}_{\! O_s}^\ast$) are one-hot embeddings of the ground truth action.

APN is trained using the same dataset as the Action Feature Extractor (AFE).
For a training sample $(s_i, s_g)$ corresponding to $[s_1, a_1, s_2, ..., s_g]$, its label is the one-hot embedding of the ground truth action $a_i$, i.e., we constrain $P = {\rm APN}(s_1, s_g)$ to maximize the probability of $a_1$.

\begin{figure}[!t]
	\centering
	\includegraphics[width=3.3in]{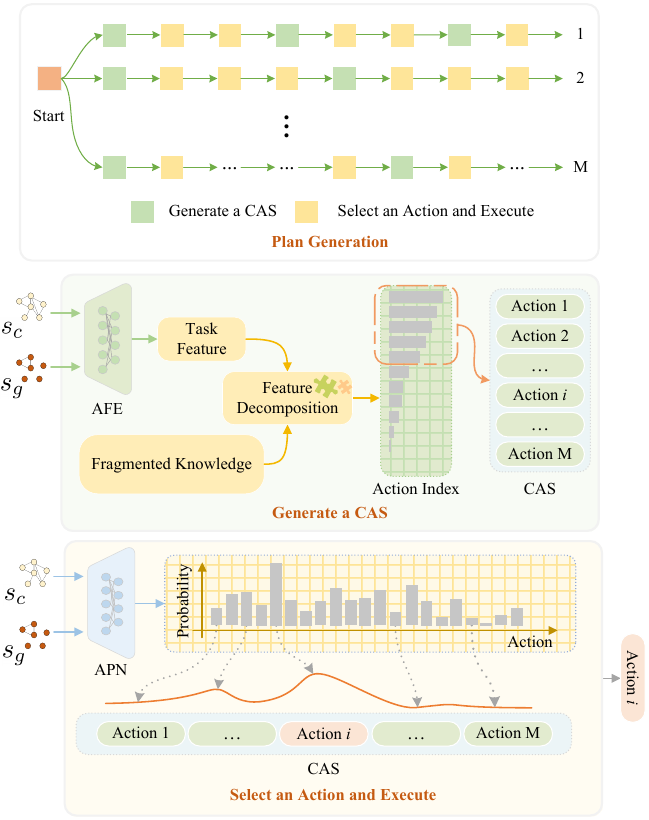}
	\caption{Schematic diagram of the plan generation.}
	\label{fig_plan_generation}
\end{figure}

\subsubsection{Plan Generation}
\label{section_plan}
The plan generation is shown in Fig. \ref{fig_plan_generation} and Algorithm \ref{algorithm_plan}.
For a specific task, the Candidate Action Generator (CAG) is used to generate CAS/MCAS. 
We take a CAS as an example to explain the plan generation. 
First, the Action Feature Extractor (AFE) is used to extract the task feature $\boldsymbol{F}_{\! T}$ based on the current state $s_c$ and the goal state $s_g$, as shown in Eq.(\ref{eq7}). 
Second, the task feature $\boldsymbol{F}_{\! T}$ is then decomposed according to the learned fragmented knowledge, as described in Eq.(\ref{eq12}). 
CAS/MCAS are generated based on the action index vector derived from the feature decomposition. 
The actions within the CAS/MCAS are unordered.
Third, APN is used to generate the probability distributions based on $s_c$ and $s_g$, as shown in Eq.(\ref{eq13}). 
After that, the action with the highest probability from each CAS is selected and executed, thus a new environmental state $s_{new}$ is reached.
APN takes $s_{new}$ as $s_c$ and repeats the previous iteration until all candidate actions have been executed.

Since a CAS may contain incorrect actions, early stopping is applied to avoid the wrong action execution, i.e., only the first several actions are executed. 
If the goal is reached, the plan generation process is terminated; otherwise, another CAS is generated by CAG based on the current and goal states. 
The task fails if 
a) the goal is still not reached after the maximum allowed step 
or 
b) some actions are not executed successfully, e.g., when the agent has something in its hand, it cannot pick up another object.

The plans for each CAS in MCAS are generated in parallel. 
From the perspective of ensemble learning, using MCAS still makes sense to improve performance.

For the whole plan generation process, we can think of a virtual environment as a human-generated scene in the brain. 
Actions are chosen in parallel from MCAS when processing a task and then executed simultaneously in multiple virtual environments. 
The generated plan is the result of brain thinking.
For practical applications, the successful plan generated from virtual environments will be executed in the real environment.
Additionally, the execution of the selected plan in the real environment can be step-by-step, i.e., after each step, it returns to virtual environments for further planning.

In addition, as some actions in a new task may not show in MCAS, i.e., never learned before, we use all executable action set $\mathbb A_{all}$ to improve generality.
The four parts of an action in $\mathbb A_{all}$ are selected in turn according to their importance, i.e., $\boldsymbol{P}_{\!\! A} \! \rightarrow \!  \boldsymbol{P}_{\! O_1} \! \rightarrow \! \boldsymbol{P}_{\! O_2} \! \rightarrow \! \boldsymbol{P}_{\! O_s}$.
The plan generated based on $\mathbb A_{all}$ is executed in parallel with the plans of MCAS.

\section{Experiments}
\label{section_4}
In this section, 
we provide the experimental results of TAL and the comparison with the baselines.
First, we introduce the experimental setup. Second, we explain the baseline settings. 
Third, we provide the experimental comparison.
After that, ablation experiments are presented. Finally, discussions are provided.

\subsection{Experimental Setup} \label{section_VI_A_setup}
\subsubsection{Environment}
We conduct the experiments with a virtual indoor scene \cite{tuli2021tango} built in the physical simulator Pybullet \cite{coumans2016pybullet}. 
The environment contains a total of 35 objects. The mobile robot consists of a robotic arm (a Universal Robotics (UR5) arm) and a mobile base (a Clearpath Husky mobile base). 
The robot can perform a total of 11 actions, such as ``pick''.

For the task-agnostic exploration, we modify the environment to improve the stability of the exploration. 
First, the robot may perform actions that are unusual in daily life during exploration, such as placing an orange on a water bottle. 
Therefore, we add constraints to avoid putting things on top of objects with uneven surfaces. 
Second, the body of the robot may affect the result of the action, for example, by preventing objects from falling. 
We set the robot to back off a certain distance after acting as a push or placement to ensure the object falls. 
Third, to ensure that the current action is performed after the simulation of the previous action has ended, we set the simulation to end only when the displacements of all objects are less than a threshold.

The action $a$ consists of an action name and one or two parameters (object name or object state).
The environmental state $s$ is represented by a scene graph, where nodes contain object information (including object name, state, size, and posture), and edges provide relationship information between objects (Close/Inside/On/Stuck). 
The environmental details are summarized in Table \ref{tab_env}, with further information available in \cite{tuli2021tango}.

\begin{algorithm}[t]
	\DontPrintSemicolon
	\SetAlgoLined
	\caption{Plan generation}
	\algorithmfootnote{$N_*$ is the number for early stopping.}
	\label{algorithm_plan}
	\KwIn {Current state $s_c$, goal state $s_g$}
	\KwOut {Plans.}
	\Init{
		$s_c = s_0$ \textcolor{gray}{// $s_0$ is the initial state of a new task.}\\
		$\boldsymbol{F}_{\!\! A}$ \textcolor{gray}{// Fragmented knowledge.}\\
		$\boldsymbol{F}_{\! T}={\rm AFE}(s_c,\ s_g)$\\
		$\{\mathbb{A}_{i},\ \mathbb{A}_{j},\ ...,\ \mathbb{A}_{k}\} = {\rm generate\_by\_CAG}(\boldsymbol{F}_{\!\! A},\ \boldsymbol{F}_{\! T})$\\
		$max\_step = 60$\\
	}
	\Parallel {
		\For {$\mathbb{A}_{*} \; in \; \{\mathbb{A}_{i},\ \mathbb{A}_{j},\ ...,\ \mathbb{A}_{k}\} $} {
			$select\_num = 0,\ plan_{*} \! = [\;],\ success={\rm False}$\\
			\For {$step=0 \;\; to \;\; max\_step$}	{
				\If {
					$select\_num\ \textgreater\ N_*$
				}{ 
					$\boldsymbol{F}_{\! T}={\rm AFE}(s_c,\ s_g)$\\
					$\mathbb{A}_{*} \! ={\rm CAG}(\boldsymbol{F}_{\!\! A},\ \boldsymbol{F}_{\! T})$\\
					$select\_num = 0$\\
				}
				$a \! $ = ${\rm select}\big(\mathbb{A}_{*} \ \vert \ {\rm APN}(s_c,\ s_g)\big)$\\
				$plan_{*} \! = plan_{*} \! + [a]$\\
				$s_c = {\rm execute}(s_c,\ a)$\\
				$+\!+\! select\_num$\\
				$success = {\rm check\_task}(s_c,\ s_g)$\\
				\If {$success$
				}{
					break\\
				} 
			} 
			$plan_{*} \! = plan_{*} \! + [success]$ \textcolor{gray}{// Add success flag.}\\
		} 
		
	}
	\Return{$\{plan_{i},\ plan_{j},\ ...,\ plan_{k}\}$}
\end{algorithm}

\subsubsection{Dataset}
Three steps are performed to construct the dataset. 
The first step is to create the action set $\mathbb A_{all}$ that contains all executable actions for environment exploration.
The second step is to perform the task-agnostic exploration to build the knowledge graph. 
The third step is to generate the dataset from the knowledge graph.

In the first step, we combine different action names, object names, and object states to generate various actions. 
We finally get an action set with a total of 3598 actions.
It is important to note that incorrect actions, such as ``changeState $[$apple$]$ $[$open$]$'', are filtered out by environmental feedback.
The remaining 1364 actions after filtering form $\mathbb A_{all}$.

In the second step, the task-agnostic exploration is performed for knowledge graph generation. 
The whole process is shown in Algorithm \ref{algorithm_explore}.
The $max\_round$ in the algorithm is set to 600.
The $max\_step$ is set to 20 for the first round and 5 for the remaining rounds.
In addition, the maximum number of attempts to explore the same node is set to 30.
Finally, a knowledge graph is built with 2995 states and 855 different actions.

In the third step, the dataset is generated from the knowledge graph. 
We sample paths (i.e., trajectories) from the knowledge graph. 
In each path, the start node corresponds to the initial environmental state, and the end node is taken as the goal state of a task.
The action sequence corresponding to the edges in each path is a solution plan.

To evaluate the performance in detail, two datasets are constructed for experiments.

\begin{table}[t]
	\centering
	\footnotesize
	\caption{Environmental details}
	\label{tab_env}
	\begin{threeparttable}
		\begin{tabular}{m{1.7cm}<{\centering} | m{6.2cm}}
			\toprule
			Action Names  (11) & pushTo, pickNplaceAonB, moveTo, drop, pick, climbUp, climbDown, clean, changeState, apply, stick\\
			\midrule
			Object Names  (35) & floor, walls, door, fridge, cupboard, table, table2, couch, book, paper, gray-cube, green-cube, red-cube, tray, tray2, big-tray, blue-bottle,  gray-bottle, red-bottle, chair, stick, box, apple, orange, dumpster, light, milk, shelf, glue, tape, stool, mop, sponge, vacuum, dirt \\
			\midrule
			Object States  (28) & Inside/Outside, On/Off, Open/Close, Up/Down, Clean/ Dirty, Grabbed/Free, Sticky/Not-Sticky, Welded/Not-Welded, Drilled/Not-Drilled, Driven/Not-Driven, Fueled/Not-Fueled, Cut/Not-Cut, Painted/Not-Painted, Same/Different-Height\\
			\bottomrule
		\end{tabular}
	\end{threeparttable}
\end{table}

\begin{figure}[t]
	\centering
	\includegraphics[width=3.3in]{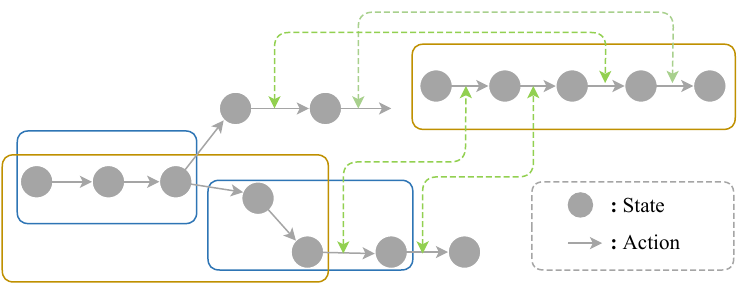}
	\caption{
		Data schematic diagram. A \textcolor{RoyalBlue}{blue} rounded rectangle represents the training data, while a \textcolor{brown}{brown} rounded rectangle is the test data. 
		A \textcolor{LimeGreen}{green} dotted line indicates the two same actions.
	}
	\label{fig_data_schematic}
\end{figure}

\textbf{Dataset-{\uppercase\expandafter{\romannumeral1}}:}
300 trajectories with the same length are sampled, and the trajectory length (the number of edges/actions in a trajectory) ranges from 1 to 10, i.e, tasks require 1 to 10 actions to be completed.
The generated dataset contains 3000 trajectories.
The training set, validation set, and test set are constructed by stratified sampling in a ratio of 6:2:2 for each length, resulting in 1800 trajectories in the training set, 600 trajectories in the validation set, and 600 trajectories in the test set.

\textbf{Dataset-{\uppercase\expandafter{\romannumeral2}}:}
In this dataset, 
\textbf{only short trajectories (with 1$\sim$3 actions) are used for fragmented knowledge learning, and long trajectories (with 4$\sim$10 actions) are used for evaluation.}
The short ones are regarded as fragmented experiences, and the long ones are regarded as new tasks.
For the training set, the trajectory length ranges from 1 to 3, and 800 trajectories are sampled for each corresponding length, resulting in 2400 trajectories.
For the validation set and the test set, the trajectory length ranges from 4 to 10, and 500 trajectories are sampled for each corresponding length, resulting in two sets containing 3500 trajectories.

For each dataset, we preferentially choose trajectories that have different endpoints and do not contain each other to increase the differences between trajectories. 
The training set is used to train AFE for fragmented knowledge learning and APN for generating the action proposal.
The validation set is used to evaluate the model training and adjust hyper-parameters for better adaptation to new tasks.
The test set is used to evaluate the performance of the proposed method and baselines on new tasks.
All tasks in the test set can be considered as new tasks, since they need new combinations of actions to reach their goal states.
The data schematic diagram is shown in Fig. \ref{fig_data_schematic}.

\subsubsection{Evaluation Metrics}
The main evaluation metric is the task success rate:
$R_{S} = N_{Success} / N_{Total}$, where $N_{Success}$ is the number of successful tasks, $N_{Total}$ is the total number of test tasks.
The criterion for judging whether the task is completed is related to the environmental state and the goal state. 
As suggested by \cite{tuli2021tango}, the criterion should satisfy some conditions: 
a) The comparison involves only task-related objects; 
b) Each task-related object must have a reference object for comparison; 
c) To consider the impact of various actions, the distance threshold for each action should be different.

Two additional evaluation metrics, the incorrect rate $R_{I}$ and the error rate $R_{E}$,  are introduced to assist in analyzing the performance of the methods.
$R_{I} = N_{Incorrect} / N_{Total}$, where $N_{Incorrect}$ represents the number of tasks that failed due to exceeding the maximum step limit.
$R_{E} = N_{Error} / N_{Total}$, where $N_{Error}$ denotes the number of tasks that failed due to error action execution, such as dropping an object before grasping it or attempting to pick up an item while already holding another.

\begin{figure}[t]
	\centering
	\includegraphics[width=3.5in]{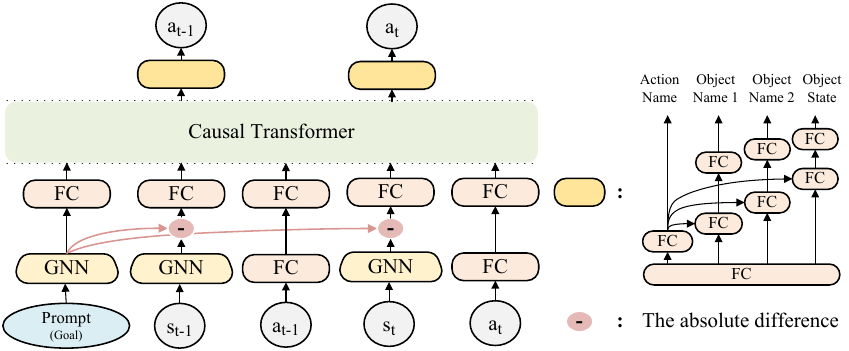}
	\caption{Architecture of Plan Transformer.}
	\label{fig_ppt_model}
\end{figure}

\subsubsection{Implementation Details}
The textual data in the state is encoded by ConceptNet \cite{speer2017conceptnet} to input to TAL.
The action feature dimension is initially set to $1 \! \times \! 4096$ and subsequently reduced to $1 \! \times \! 500$ using PCA.
The setting of the Multiple Candidate Action Sets (MCAS) is heuristic. 
We set it to contain 7 action sets. 
The action numbers and the number for early stopping in each set are [(5, 2), (10, 5), (15, 5), (20, 5), (20, 10), (30, 5), (30, 10)]. 
The maximum number of steps allowed per task is set to 60. 
3 random seeds are used in the experiments.

\subsection{Baselines}
We compare TAL with 
the offline model-free RL method CQL \cite{kumar2020conservative},
the model-based RL method TD-MPC2 \cite{tdmpc2},
the offline meta-RL method Meta-DT \cite{metadt},
the IL method Plan Transformer (PT), which is designed based on the transformer \cite{radford2018improving},
and the IL method BC.

The IL methods here are slightly different from typical IL methods.
For typical IL, the training data comes from experts, but the training set is treated as expert data in our setting.
We aim to explore decision-making and learning methods that minimize human involvement, in other words, we have no experts but only data obtained through task-agnostic exploration.

\subsubsection{CQL}
The network of CQL is the same as the Action Proposal Network (APN) (Section \ref{section_3_APN}).
According to the action’s representation, 
The output consists of four Q-tables corresponding to action name, object name 1, object name 2, and object state.
The policy is trained using the same training set as APN and the Action Feature Extractor (AFE).
In other words, the training set is taken as a fixed replay buffer.

All tasks are unified into a consistent form, which is generating action sequences that minimize the gap between the environmental state and the goal. 
We set the reward function by the Manhattan distance between the environmental state and the goal.
\begin{equation}\label{eq16}
\begin{aligned}
{\rm R_c}\left(s_{c}, s_{g}\right) \! = \!
\left\{
\begin{array}{lr}
\!\! 100, \: if \: done\ (task\ completed)\\
\!\! 1, \: \: \: \: \: if \: {\rm L_1}(s_{c}, s_{g})  <  d\\
\!\! 0, \: \: \: \: \: Others.
\end{array}
\right.\\
\end{aligned}
\end{equation}
Here, $d$ is the shortest distance between all previous states and the goal state. 
$s_{c}$ and $s_{g}$ are current and goal states.

\subsubsection{TD-MPC2}
The model-based RL method TD-MPC2 combines the advantages of Temporal Difference learning and Model Predictive Control.
To make the model fit our datasets and converge, the following settings are adopted:
a) The action output consists of four parts, which are consistent with CQL.
b) For the policy prior of the world model, i.e., the network used to generate the Gaussian distribution with mean and std, we constrain the mean to be close to the ground truth action.
c) The policy prior loss is summed with other losses and back-propagated together, rather than optimizing each part separately as in the original setting.

\begin{figure*}[!t]
	\centering
	\includegraphics[width=.95\linewidth]{./src/experiment_results_v14_900.png}
	\caption{
		Performance comparison ($R_{S}$) on Dataset-{\uppercase\expandafter{\romannumeral1}}, which consists of 10 subsets, each requiring a different number of actions to complete.
	}
	\label{fig_results}
\end{figure*}

\begin{table}[t]
	\centering
	\footnotesize
	\renewcommand{\arraystretch}{1.6}
	\setlength\aboverulesep{0pt}\setlength\belowrulesep{0pt}
	\setcellgapes{3pt}\makegapedcells
	\tabcolsep=3.0pt
	\caption{Performance comparison on Dataset-{\romannumeral1}.}
	\label{tab_performance_1_avg}
	\begin{threeparttable}
		\begin{tabular}{c | *{7}{c}}
			\toprule
			Method & CQL & TD-MPC2 & TD-MPC2$^\ddag$ & Meta-DT & PT & BC & TAL \\
			\midrule
			$\bm{R_{S}} \uparrow$ & 10.50 & $\,$ 2.33 & $\,$ 9.33 & $\,$ 5.33 & $\,$ 8.67  & 23.56 & \textbf{45.78} \\
			$R_I \ \: \mbox{-}$   & 71.67 & $\,$ 0.00 & 66.50     & $\,$ 0.00 & 88.00      & 58.67 & 15.72 \\
			$R_E \downarrow$      & 17.83 & 97.67     & 24.17     & 94.67     & $\,$ 3.33  & 17.78 & 38.50 \\
			\bottomrule
		\end{tabular}
		\begin{tablenotes}[flushleft]
			\scriptsize
			\item
			$\ddag$:  TD-MPC2 (w/o MPC).
		\end{tablenotes}
	\end{threeparttable}
\end{table}

\subsubsection{Meta-DT}
The offline meta-RL method Meta-DT consists of a context-aware world model and a meta-decision transformer. 
The output of the context model contains task-relevant information and will be input to the transformer.
The zero-shot setting is used in the experiments, i.e., without the prompt component.
To make the model fit our datasets and converge, we make the following settings:
a) The action output consists of four parts, same as CQL and TD-MPC2.
b) The goal state is input to the context model to encode contextual information.
Specifically, for the context encoder: $E_{\psi}(z_t^i|{\mu}_t^i, s_g)$, for reward decoder: $R_{\phi}(r_t|s_t, a_t, z_t^i, s_g)$.
c) Since the reward function Eq.(\ref{eq16}) only contains 3 values, the reward decoder is trained in a classification manner.
d) The goal state is input to the transformer to predict actions.
e) More powerful models are used. 
ResNet-18 is used for the context encoder and decoder. 
The Transformer is set to 6 layers, 4 heads, and a hidden dimension of 1024.

\subsubsection{PT}
PT is based on the transformer architecture \cite{radford2018improving}, as shown in Fig. \ref{fig_ppt_model}.
The inputs of PT are prompt (goal state), states, and actions in history. 
The output is an action for current execution. 
We feed the goal state to the network as a prompt so that PT can perform various tasks. 
We set the maximum trajectory length of the model input to 5.
The GNN structure and settings are identical to those of APN.
The loss function is the cross-entropy loss.

\subsubsection{BC}
As shown in Section \ref{section_3_APN}, APN takes the current environmental state $s_c$ and the goal state $s_g$ as input and outputs the probability distributions of all executable actions.
If we select actions directly based on the output of APN instead of selecting from MCAS, then APN can be used as a BC method.
According to the output of APN, the four parts of an action are selected in turn according to their importance, i.e., $\boldsymbol{P}_{\!\! A} \! \rightarrow \!  \boldsymbol{P}_{\! O_1} \! \rightarrow \! \boldsymbol{P}_{\! O_2} \! \rightarrow \! \boldsymbol{P}_{\! O_s}$.

\begin{table*}[t]
	\centering
	\footnotesize
	\renewcommand{\arraystretch}{1.6}
	\setlength\aboverulesep{0pt}\setlength\belowrulesep{0pt}
	\setcellgapes{3pt}\makegapedcells
	\tabcolsep=4.2pt
	\caption{Ablation experiment results on Dataset-{\romannumeral1}.}
	\label{tab_ablation_1}
	\begin{threeparttable}
		\begin{tabular}{*{7}{c} | *{3}{c} | *{10}{c}}
			\toprule
			\multicolumn{7}{c |}{Setting} & \multicolumn{3}{c |}{Average} & \multicolumn{10}{c}{Number of Actions*}\\
			\cline{1-20}
			{\scriptsize AFE} & {\scriptsize GNN} & {\scriptsize APN} & {\scriptsize MCAS} & {\scriptsize ES} & {\scriptsize PCA}& {\scriptsize $\mathbb A_{all}$} & $\bm{R_{S}} \uparrow$ & $R_{I}$ & $R_{E} \downarrow$ & 1 & 2 & 3 & 4 & 5 & 6 & 7 & 8 & 9 & 10 \\
			\midrule

			\ding{55} & \usym{2713} &  \usym{2713} & \usym{2713} & \usym{2713} & \usym{2713} &  \usym{2713} & 40.17 & 15.67 & 44.17 & 93.33 & \textbf{81.67} & 68.33 & 46.67 & 33.33 & 28.33 & 20.00 & \textbf{20.00} & $\,$ 3.33 & $\,$ 6.67 \\

			\usym{2713} & \ding{55} &  \usym{2713} & \usym{2713} & \usym{2713} & \usym{2713} &  \usym{2713} & 39.50  & 11.50 & 49.00 & 91.11 & 75.56 & 66.67 & 50.56 & 33.89 & \textbf{37.22} & 15.56 & 14.44 & $\,$ 4.44 & $\,$ 5.56 \\

			\usym{2713} & \usym{2713} &  \ding{55} & \usym{2713} & \usym{2713} & \usym{2713} &  \usym{2713} & 32.17  & 26.50 & 41.33 & 78.89 & 65.00 & 57.78 & 38.89 & 29.44 & 22.22 & 13.89 & $\,$ 6.67 & $\,$ 1.67 & $\,$ 7.22 \\

			\usym{2713} & \usym{2713} &  \usym{2713} & \ding{55} & \usym{2713} & \usym{2713} & \usym{2713} & 33.00  & 22.94 & 44.06 & 93.33 & 72.22 & 56.67 & 38.33 & 31.67 & 15.56 & 10.00 & $\,$ 7.22 & $\,$ 0.56 & $\,$ 4.44 \\

			\usym{2713} & \usym{2713} &  \usym{2713} & \usym{2713} & \ding{55} & \usym{2713} & \usym{2713} & 37.06  & $\,$ 2.61 & 60.33 & 93.33 & 79.44 & 58.89 & 47.78 & 36.67 & 19.44 & 17.22 & 10.00 & $\,$ 1.11 & $\,$ 6.67 \\

			\usym{2713} & \usym{2713} &  \usym{2713} & \usym{2713} & \usym{2713} & \ding{55} & \usym{2713} & 36.83  & $\,$ 6.00 & 57.17 & \textbf{98.33} & 79.44 & 63.89  & 38.89 & 31.67 & 19.44 & 11.67 & $\,$ 9.44 & $\,$ 5.00 & \textbf{10.56} \\

			\usym{2713} & \usym{2713} &  \usym{2713} & \usym{2713} & \usym{2713} & \usym{2713} & \ding{55} & 41.78  & 15.00 & 43.22 & 93.89 & 80.00 & 68.33 & \textbf{58.89} & 39.44 & 25.00 & 20.00 & 15.00 & $\,$ 7.22 & 10.00 \\	

			\usym{2713} & \usym{2713} &  \usym{2713} & \usym{2713} & \usym{2713} & \usym{2713} & \usym{2713} & \textbf{45.78}  & 15.72 & 38.50 & \textbf{98.33} & \textbf{81.67} & \textbf{71.11}  & \textbf{58.89} & \textbf{51.11} & 31.67 & \textbf{25.00} & 17.78 & \textbf{11.67} & \textbf{10.56} \\

			\bottomrule
		\end{tabular}
		\begin{tablenotes}[flushleft]
			\scriptsize
			\item
			*: ``Number of Actions'' is the number of actions in the ground truth plan. 
			Bold indicates the best result.
		\end{tablenotes}
	\end{threeparttable}
\end{table*}

\begin{table}[t]
	\centering
	\scriptsize
	\renewcommand{\arraystretch}{1.6}
	\setlength\aboverulesep{0pt}\setlength\belowrulesep{0pt}
	\setcellgapes{3pt}\makegapedcells
	\tabcolsep=2.3pt
	\caption{Performance comparison on Dataset-{\romannumeral2}.}
	\label{tab_performance_2}
	\begin{threeparttable}
		\begin{tabular}{c | *{3}{c} | *{7}{c}}
			\toprule
			\multirow{2}*{Method} & \multicolumn{3}{c |}{Average} & \multicolumn{7}{c}{Number of Actions*}\\
			\cline{2-11}
			& $\bm{R_{S}} \uparrow$ & $R_{I}$ & $R_{E} \downarrow$ & 4 & 5 & 6 & 7 & 8 & 9 & 10\\
			\midrule
			CQL                           & $\,$ 3.17 & 74.57     & 22.26 &  $\,$ 7.60 & $\,$ 5.20 & $\,$ 5.20 & $\,$ 1.00 & $\,$ 2.60 & $\,$ 0.40 & 0.20 \\
			TD-MPC2                   & $\,$ 0.60 & $\,$ 0.00 & 99.40 &  $\,$ 1.20 & $\,$ 0.80 & $\,$ 1.20 & $\,$ 0.40 & $\,$ 0.60 & $\,$ 0.00 & 0.00 \\
			TD-MPC2$^\ddag$     & $\,$ 3.86 & 85.14     & 11.00 & 10.60      & $\,$ 7.20 & $\,$ 4.60 & $\,$ 2.00 & $\,$ 1.80 & $\,$ 0.80 & 0.00 \\
			Meta-DT                     & $\,$ 4.71 & $\,$ 3.57 & 91.71 & 11.80      & $\,$ 7.40 & $\,$ 6.00 & $\,$ 3.00 & $\,$ 3.00 & $\,$ 1.60 & 0.20 \\
			PT                              & $\,$ 3.06 & 86.66     & 10.29 &  $\,$ 6.80 & $\,$ 7.00 & $\,$ 5.60 & $\,$ 4.60 & $\,$ 2.20 & $\,$ 1.40 & 0.40 \\
			BC                              & $\,$ 6.80 & 76.79     & 16.41 & 19.60      & 11.67     & $\,$ 7.27 & $\,$ 4.33 & $\,$ 2.93 & $\,$ 1.27 & 0.53 \\
			TAL (Ours)          & \textbf{29.96}        & 27.90 & 42.13      & \textbf{59.67} & \textbf{47.07} & \textbf{33.87} & \textbf{28.27} & \textbf{19.13} & \textbf{13.20} & \textbf{8.53}\\
			\bottomrule
		\end{tabular}
		\begin{tablenotes}[flushleft]
			\scriptsize
			\item
			*: ``Number of Actions'' is the number of actions in the ground truth plan. Bold indicates the best result.
			$\ddag$:  TD-MPC2 (w/o MPC).
		\end{tablenotes}
	\end{threeparttable}
\end{table}

\subsection{Performance Comparison}

\subsubsection{Dataset-{\uppercase\expandafter{\romannumeral1}}}
The average success rates of compared methods on the entire test set are provided in Table \ref{tab_performance_1_avg}. 
As can be seen, our TAL produces the highest success rate of 45.78\%, 
which outperforms other methods by more than 20\%.

For a more detailed comparison,
we split the test set into 10 subsets according to the number of actions required to complete tasks. 
The success rates of compared methods on each test subset are shown in Fig. \ref{fig_results}. 
TAL achieves the highest average success rate on every test subset. 
The success rate of all models gradually decreases when the number of actions increases.
PT and CQL have significantly lower success rates than TAL on each subset.
For example, on task subset 2, TAL achieves a success rate of 81.67\%, while BC, CQL, and PT drop to 52.22\%, 26.67\%, and 20.00\%, respectively.
On task subset 5, TAL has a success rate of 51.11\%, while the success rates of BC, CQL, and PT are 38.38\%, 13.33\%, and 6.67\%, respectively.
On the subset 7, TAL still has a 25\% success rate, while the success rates of all other methods are  less than 10\%
The performance of TD-MPC2 is not well, achieving only 2.33\% success rate, but without MPC, the accuracy is improved by 7\%.
In addition, according to $R_{I}$ and $R_{E}$, TD-MPC2 is more likely to fail due to error action execution, whereas failure due to exceeding the step limit is more common without MPC.
Meta-DT also performed poorly, with an accuracy of 5.33\%.
The results of these two world model-based methods show that it is challenging to train world models using small datasets as in the experiments.

\subsubsection{Dataset-{\uppercase\expandafter{\romannumeral2}}}
The performance comparisons are listed in Table \ref{tab_performance_2}. 
As can be seen, TAL produces the highest success rate of 29.96\%, 
which outperforms CQL by 26.79\%, TD-MPC2 by 29.36\%, Meta-DT by 26.1\%, PT by 26.9\%, and BC by 23.16\%.
TAL also achieves the best performance on every subset.
However, compared to the performance on Dataset-{\uppercase\expandafter{\romannumeral1}}, the performance of all methods on Dataset-{\uppercase\expandafter{\romannumeral2}} dropped, which is caused by the larger distribution difference between the training set and the test set in Dataset-{\uppercase\expandafter{\romannumeral2}}.

\subsubsection{Qualitative results}
\begin{figure}[t]
	\centering
	\footnotesize
	\subfloat[]{\includegraphics[width=1.6in]{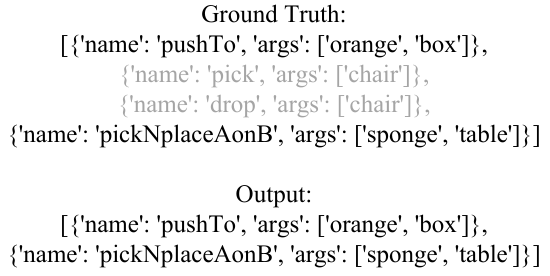}%
		\label{fig_first_case}}
	\hfil
	\subfloat[]{\includegraphics[width=1.8in]{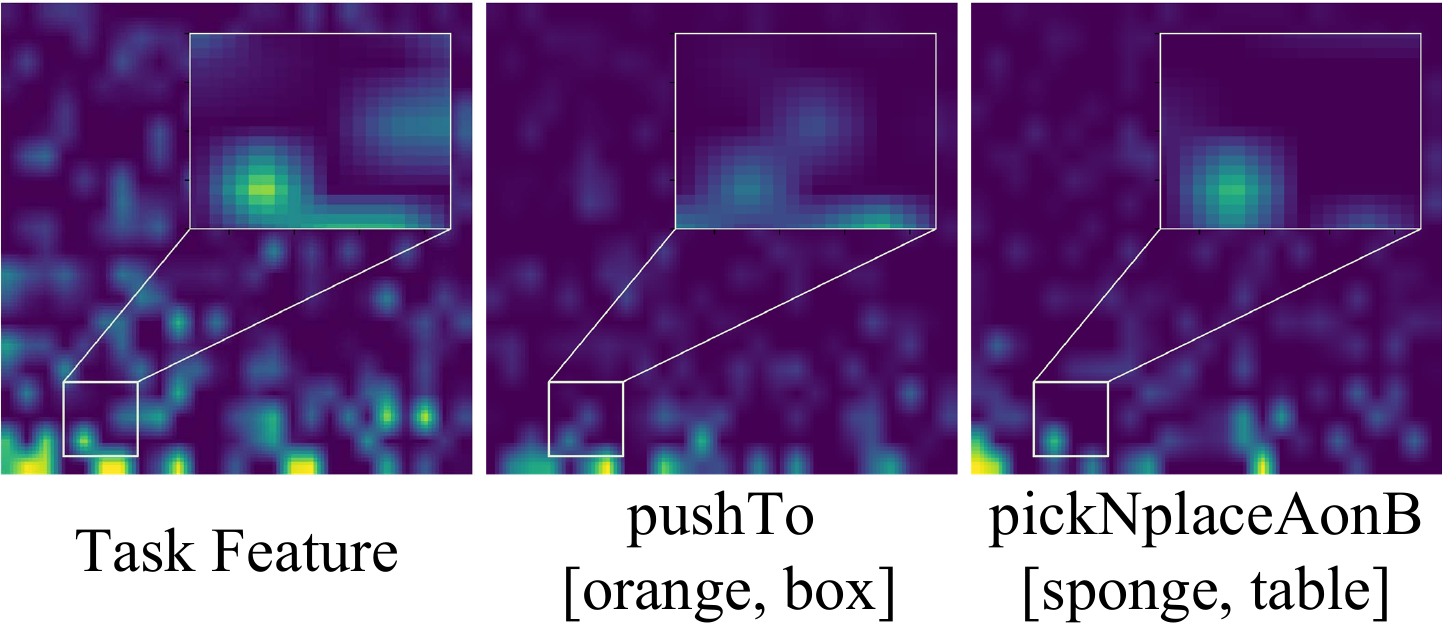}%
		\label{fig_second_case}}
	\vfil
	\vspace{-2mm}
	\subfloat[]{\includegraphics[width=3.5in]{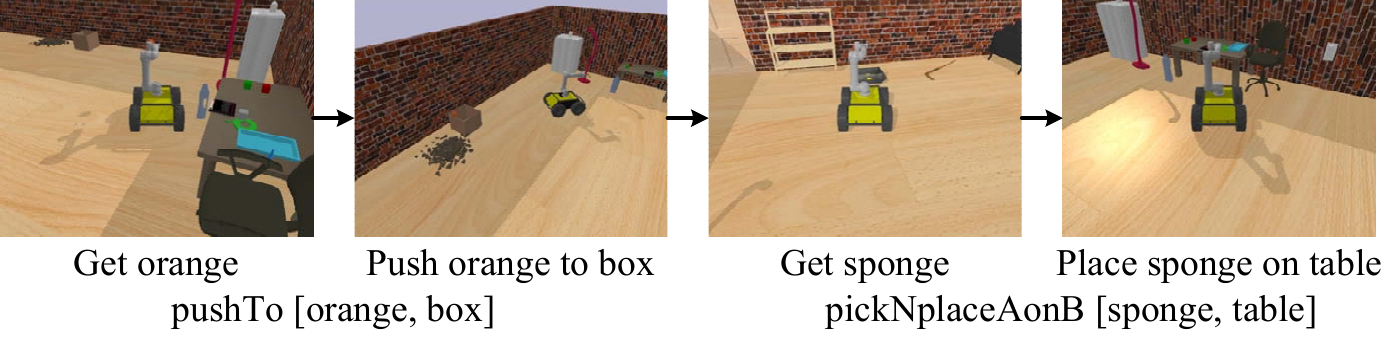}%
		\label{fig_third_case}}
	\caption{An example task. 
		(a) Ground Truth and output of our model. 
		(b) Action feature visualization.
		(c) Execute the plan in the environment.
	}
	\label{fig_qualitative}
\end{figure}

\begin{figure*}[t] 
	\centering
	\includegraphics[width=.90\linewidth]{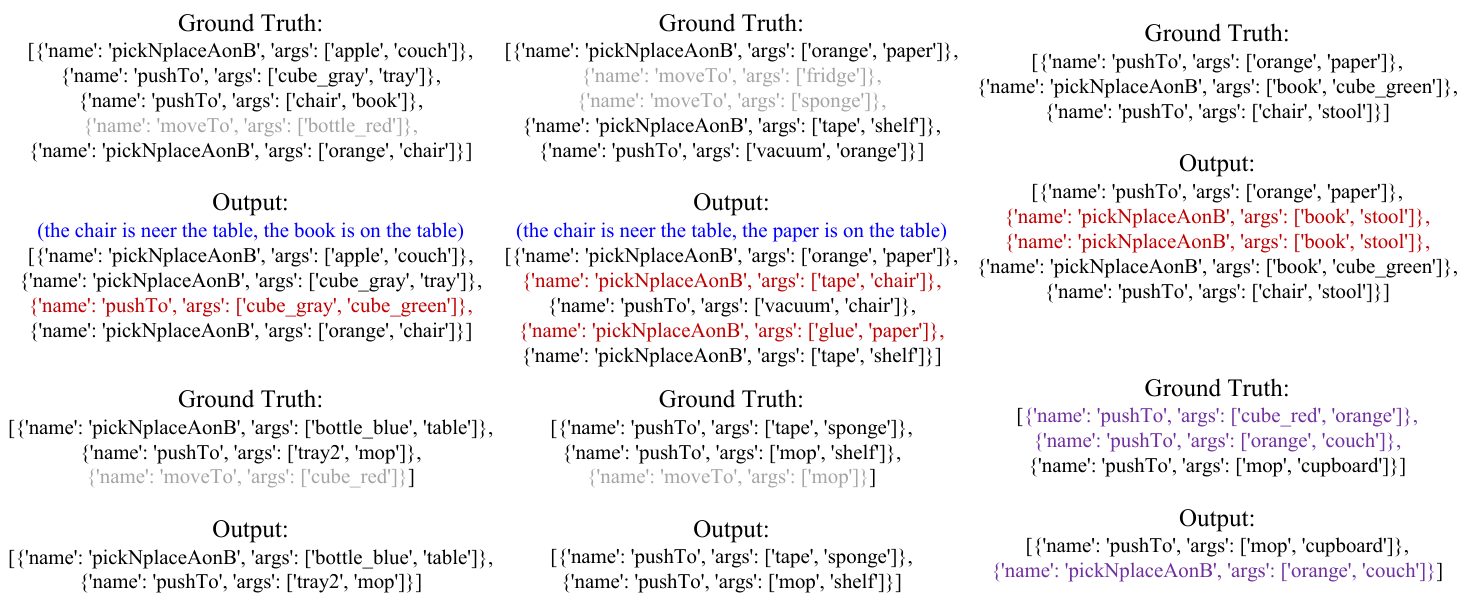}
	\caption{Qualitative results.  \textcolor{RoyalBlue}{Blue} is the description of the environment. \textcolor{gray}{Gray} indicates insignificant operations, i.e., no changes to the environment. \textcolor{red}{Red} indicates redundant actions.\textcolor{violet}{Violet} indicates knock-on impacts, where consecutive operations involve the same objects.}
	\label{fig_qualitative_text}
\end{figure*}

Fig. \ref{fig_qualitative} shows an example task and Fig. \ref{fig_qualitative_text} displays some qualitative results. 
There are four situations involved in Fig. \ref{fig_qualitative_text}: 
a) The predicted plan has some incorrect actions; 
b) The ground truth plan contains invalid actions, such as just moving around without changing the state of other objects; 
c) The states of some objects already meet the task requirements; 
d) A knock-on impact occurs in the ground truth plan, i.e., multiple different actions involve the same object. 
It can be seen that TAL performs well on these new tasks.

\subsection{Ablation Studies}
A total of seven ablation experiments are conducted to evaluate the contribution of each component, 
as shown in Table \ref{tab_ablation_1} and Table \ref{tab_ablation_2}.

\begin{table*}[t]
	\centering
	\footnotesize
	\renewcommand{\arraystretch}{1.6}
	\setlength\aboverulesep{0pt}\setlength\belowrulesep{0pt}
	\setcellgapes{3pt}\makegapedcells
	\tabcolsep=6.0pt
	\caption{Ablation experiment results on Dataset-{\romannumeral2}.}
	\label{tab_ablation_2}
	\begin{threeparttable}
		\begin{tabular}{*{7}{c} | *{3}{c} | *{7}{c}}
			\toprule
			\multicolumn{7}{c |}{Setting} & \multicolumn{3}{c |}{Average} & \multicolumn{7}{c}{Number of Actions*}\\
			\cline{1-17}
			{\scriptsize AFE} & {\scriptsize GNN} & {\scriptsize APN} & {\scriptsize MCAS} & {\scriptsize ES} & {\scriptsize PCA} & {\scriptsize $\mathbb A_{all}$} & $\bm{R_{S}} \uparrow$ & $R_{I}$ & $R_{E} \downarrow$  & 4 & 5 & 6 & 7 & 8 & 9 & 10\\
			\midrule
			
			\ding{55} & \usym{2713} &  \usym{2713} & \usym{2713} & \usym{2713} & \usym{2713} &  \usym{2713} & 23.51 & 26.72 & 49.76 & 49.07 & 37.07 & 27.80 & 21.47 & 13.07 & $\,$ 9.73 & 6.40 \\
			
			\usym{2713} & \ding{55} &  \usym{2713} & \usym{2713} & \usym{2713} & \usym{2713} &  \usym{2713} & 28.73 & 27.73 & 43.53 & 56.33 & 43.60 & 31.80 & 27.60 & 18.87 & \textbf{13.93} & \textbf{9.00} \\
			
			\usym{2713} & \usym{2713} & \ding{55} & \usym{2713} & \usym{2713} & \usym{2713} & \usym{2713}   & 12.83 & 20.20 & 66.97 & 29.73 & 19.67 & 14.07 & 11.07 & $\,$ 7.60 & $\,$ 4.40 & 3.27 \\
			
			\usym{2713} & \usym{2713} & \usym{2713} & \ding{55} & \usym{2713} & \usym{2713}  & \usym{2713}  & 17.61 & 37.01 & 45.38 & 42.00 & 28.73 & 19.27 & 14.73 & $\,$ 9.67 & $\,$ 5.27 & 3.60 \\
			
			\usym{2713} & \usym{2713} & \usym{2713} & \usym{2713} & \ding{55} & \usym{2713}  & \usym{2713}  & 26.80 & 10.87 & 62.33 & 56.73 & 40.67 & 30.00 & 23.87 & 16.87 & 11.73 & 7.73 \\
			
			\usym{2713} & \usym{2713} & \usym{2713} & \usym{2713} & \usym{2713} & \ding{55} & \usym{2713}   & 22.07 & 18.35 & 59.58 & 48.73 & 34.40 & 25.27 & 19.00 & 12.53 & $\,$ 8.60 & 5.93 \\
			
			\usym{2713} & \usym{2713} & \usym{2713} & \usym{2713} & \usym{2713} & \usym{2713} & \ding{55}   & 28.53 & 31.16 & 43.30 & 57.40 & 44.27 & 33.13 & 26.47 & 18.73 & 12.27 & 7.47 \\
			
			\usym{2713} & \usym{2713} & \usym{2713} & \usym{2713} & \usym{2713} & \usym{2713} & \usym{2713} & \textbf{29.96} & 27.80 & 42.13 & \textbf{59.67} & \textbf{47.07} & \textbf{33.87} & \textbf{28.27} & \textbf{19.13} & 13.20 & 8.53 \\
			
			\bottomrule
		\end{tabular}
		\begin{tablenotes}[flushleft]
			\scriptsize
			\item
			*: ``Number of Actions'' is the number of actions in the ground truth plan. 
			Bold indicates the best result.
		\end{tablenotes}
	\end{threeparttable}
\end{table*}

\begin{figure}[t]
	\centering
	\footnotesize
	\subfloat[]{\includegraphics[width=2.9in]{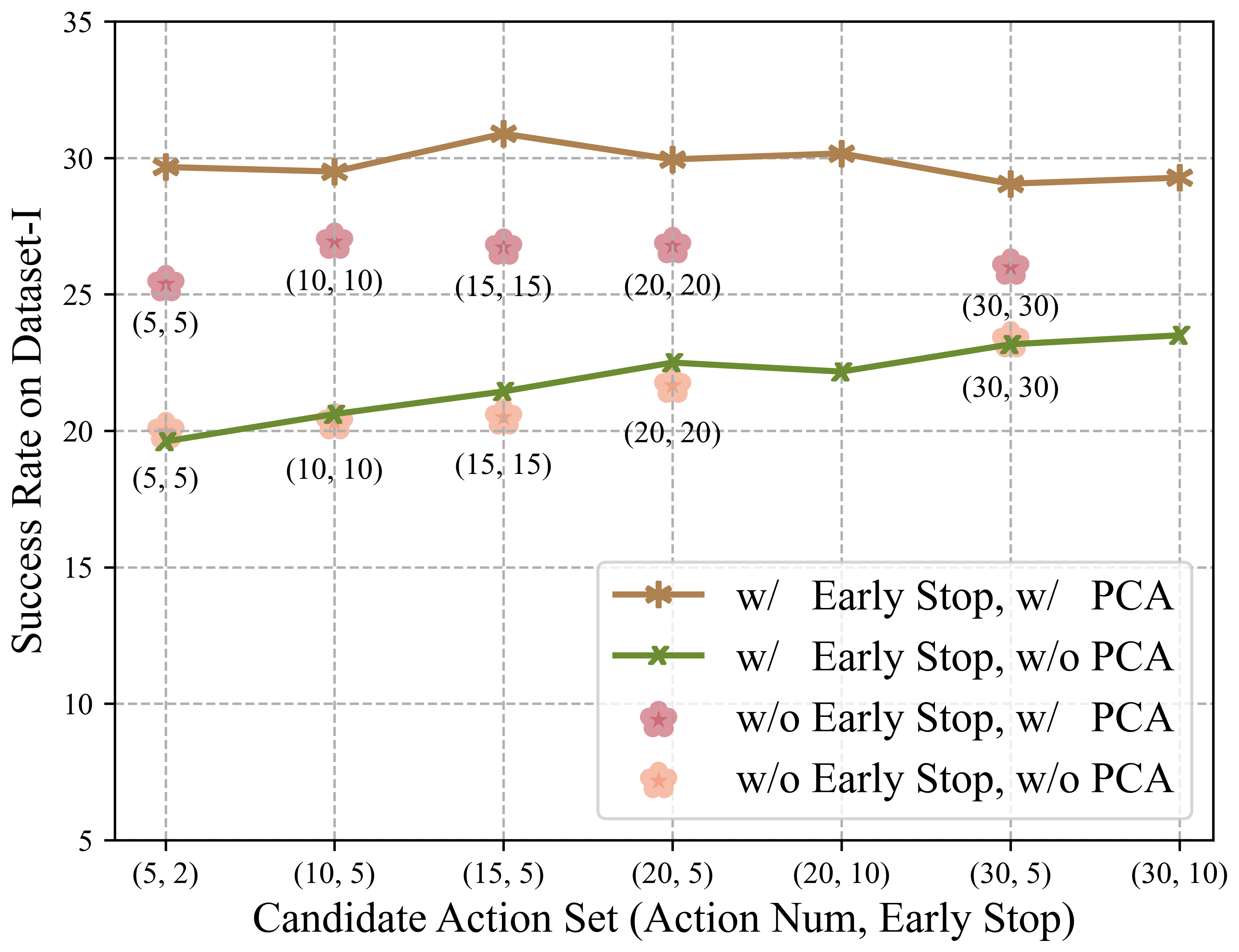}}
	\vspace{-4ex}
	\subfloat[]{\includegraphics[width=2.9in]{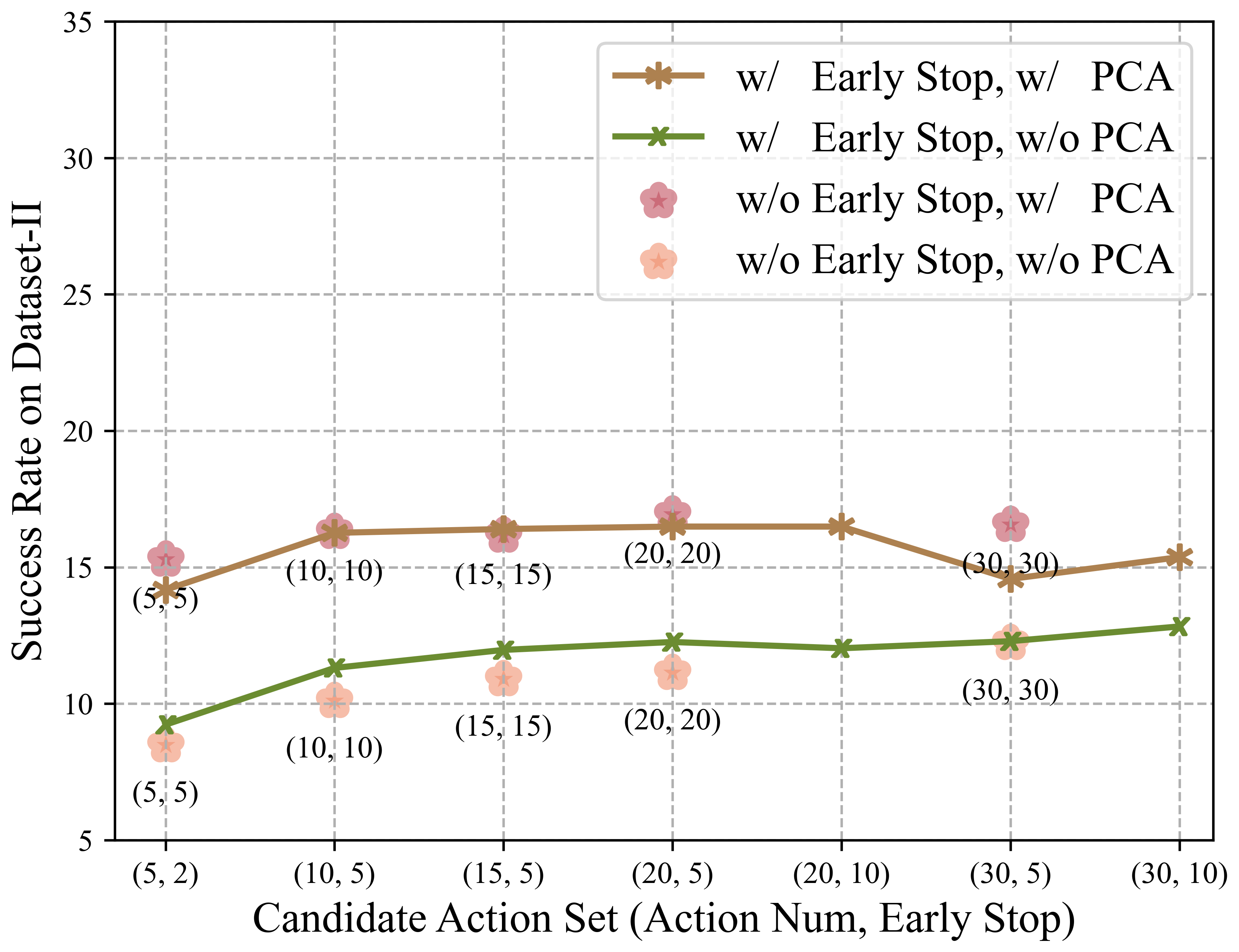}}
	\caption{Candidate action set analysis.}
	\label{cas_analysis}
\end{figure}

\begin{itemize}
	\item AFE:
	A feature extractor is trained using only the classification loss $\mathcal{L}_{cls}$ (Eq.(\ref{eq4})), and the results are regarded as without AFE to verify its effectiveness.
	\item GNN:
	To evaluate the contribution of GNN to feature extraction, we conduct an experiment replacing GNN with MLP.
	\item APN: 
	This experiment evaluates the role of the action execution order of the Multiple Candidate Action Sets (MCAS).
	If APN is used, actions in MCAS are executed according to the probability distribution generated by APN.
	Otherwise, actions in MCAS are executed sequentially according to the action index ${\widetilde{A}}_I$ (Eq.(\ref{eq12})). 
	\item MCAS:
	To verify the effectiveness of MCAS, we conduct separate experiments for each CAS within MCAS and report the best result.
	\item Early Stopping (ES):
	Corresponding to the settings of MCAS described in Section \ref{section_VI_A_setup},
	the MCAS setting here is [(2, \textbf{2}), (5, \textbf{5}), (10, \textbf{10}), (15, \textbf{15}), (20, \textbf{20}), (30, \textbf{30})], indicating without ES.
	\item Ablation experiments are conducted on PCA and $\mathbb A_{all}$, providing further insights into their roles.
\end{itemize}

\subsubsection{Dataset-{\uppercase\expandafter{\romannumeral1}}}
From Table \ref{tab_ablation_1}, we can see that APN and MCAS play the most important role. 
Using APN and MCAS results in 13.61\% and 12.78\% improvements in the average success rate, respectively. 
ES and PCA improve the average success rate by 8.72\% and 8.95\%, respectively.
AFE achieves an improvement of 5.61\%, while GNN improves by 6.28\%.
$\mathbb A_{all}$ contributes 4\%.

\subsubsection{Dataset-{\uppercase\expandafter{\romannumeral2}}}
As shown in Table \ref{tab_ablation_2}, APN and MCAS still perform well, improving performance by 17.13\% and 12.35\% respectively.
PCA improves performance by 7.89\%, followed by AFE with 6.45\% and ES with 3.16\%. 
Meanwhile, $\mathbb A_{all}$ and GNN provide smaller gains of 1.43\% and 1.23\%, respectively.

\subsection{Candidate Action Set Analysis}
Fig. \ref{cas_analysis} is an illustration to show the effectiveness of early stopping and PCA. 
In the line part, early stopping is applied, while the scatter part represents that it is not applied. 
In the case of not using PCA, the application of early stopping slightly improves success rates on both datasets. 
In the case of using PCA, early stopping significantly improves the success rate on Dataset-\uppercase\expandafter{\romannumeral1}, but not obvious on Dataset-\uppercase\expandafter{\romannumeral2}. 
Besides, whether with or without early stopping, PCA increases success rates on both datasets.

\begin{figure}[t]
	\centering
	\includegraphics[width=3.4in]{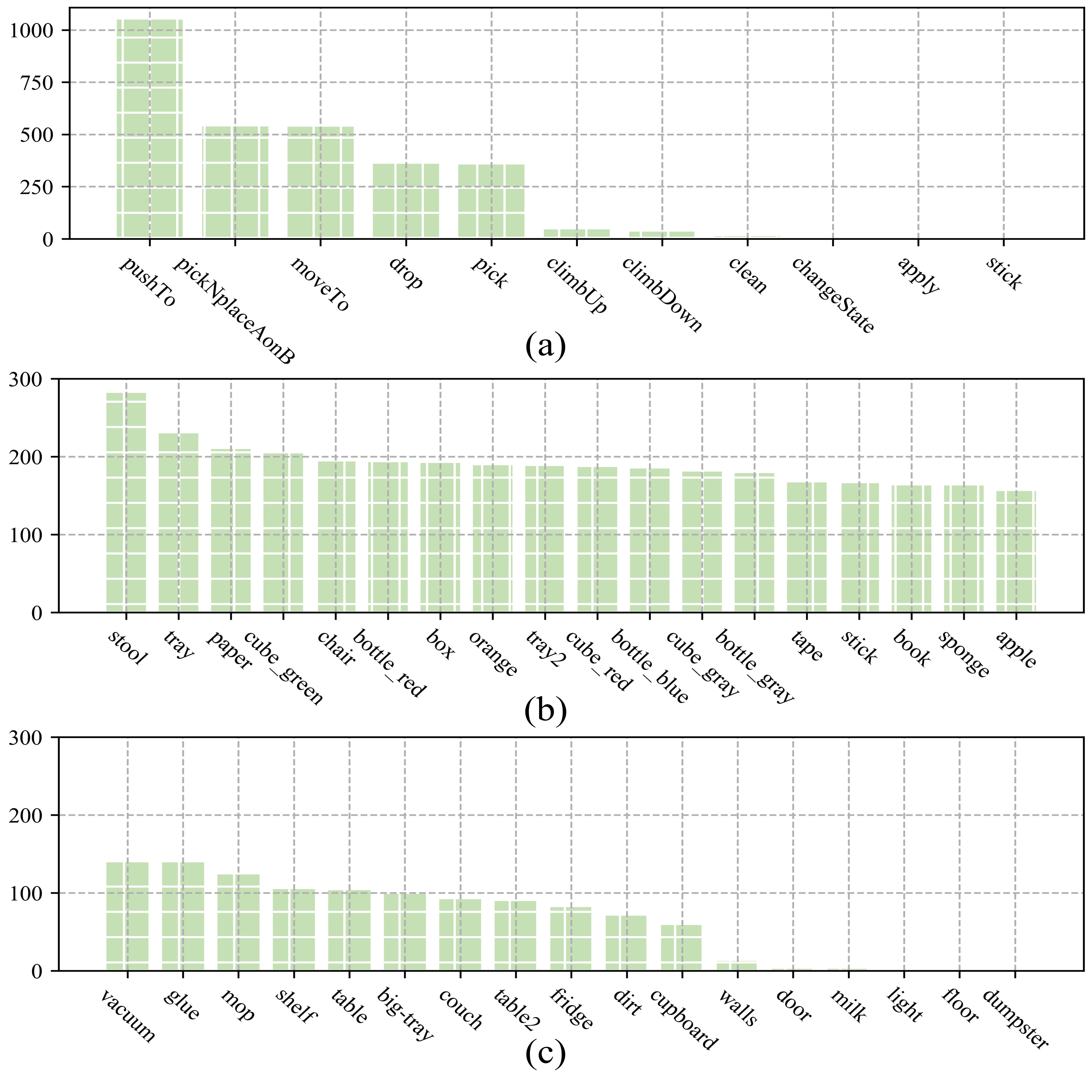}
	\caption{The number of actions and object interactions performed in the knowledge graph. 
		(a) Action execution times. 
		(b) Object Interactions (Part \uppercase\expandafter{\romannumeral1}).
		(c) Object Interactions (Part \uppercase\expandafter{\romannumeral2}).
	}
	\label{fig_dataset}
\end{figure}

\subsection{Failure Case Analysis}
Many tasks fail because the task requires actions that the agent has not done before.
In addition, we find TAL is better at solving placement tasks than state-changing tasks.
These problems are related to how the environment is set up and how it is explored. 
Almost all objects in the scene can be paired with placement-related actions, but only a few objects can be combined with state-changing actions, as shown in Fig. \ref{fig_dataset}.
Investigating more efficient exploration methods can be beneficial to improve performance.

\subsection{Limitations and Future Works}
The proposed method has some limitations. 
1) The random task-agnostic exploration method is unsafe in the real world and will lead to low-quality exploration data. 
2) The task representation using state vectors cannot meet the requirements of interactive tasks and limits the application scenarios.

Further works will be carried out in the following aspects.
1) Environment exploration:
In real-world applications, safe and efficient task-agnostic (or reward-free, intrinsic reward) exploration methods are important and need in-depth research.
2) Task representation:
Compared with the vector representation, the text representation is more advantageous in interactive tasks and can be combined with large language models to improve the performance of intelligent agents.
3) Knowledge representation:
Different actions may be combined into a meta-action, or new actions need to be learned for specific objects. 
Therefore, the representation of knowledge still needs further research.

\section{Conclusion}
\label{section_5}
In this paper, we propose a task-agnostic learning method (TAL for short) that can learn fragmented knowledge from task-agnostic data to accomplish new tasks. 
TAL consists of four stages. 
First, the task-agnostic exploration is performed to collect data, which is organized via a knowledge graph. 
Second, the action feature extractor is proposed and trained using the collected knowledge graph data for task-agnostic fragmented knowledge learning. 
Third, the candidate action generator is proposed, which applies the action feature extractor on a new task to generate multiple candidate action sets. 
Finally, the plan generation is performed based on an action proposal network. 
The experiments have confirmed the effectiveness of the proposed method on a virtual indoor scene.

\bibliographystyle{IEEEtran}
\bibliography{./src/reference}

\end{document}